 \documentclass[pmlr,twocolumn]{jmlr} 

\usepackage{booktabs}
\usepackage[load-configurations=version-1]{siunitx} 

\theorembodyfont{\upshape}
\theoremheaderfont{\scshape}
\theorempostheader{:}
\theoremsep{\newline}

\usepackage[utf8]{inputenc} 
\usepackage[T1]{fontenc}    
\usepackage{hyperref}       
\usepackage{url}            
\usepackage{booktabs}       
\usepackage{amsfonts}       
\usepackage{nicefrac}       
\usepackage{microtype}      
\usepackage{multirow}
\usepackage{multicol}
\usepackage{todonotes}
\usepackage{amsbsy}

\newcommand{\PT}{\mathrm{PT}}

\newcommand{\NO}{\mathrm{noiser}}

\renewcommand{\vec}{\boldsymbol}

\jmlrvolume{}
\jmlryear{}
\jmlrsubmitted{}
\jmlrpublished{}
\jmlrworkshop{} 

\title{Adversarial Contrastive Pre-training for Protein Sequences}

\newcommand{\rr}{\boldsymbol}

\author{%
\Name{Matthew B.A. McDermott} \Email{mmd@mit.edu}\\
\Name{Brendan Yap} \Email{yap@mit.edu}\\
\Name{Tzu Ming Harry Hsu} \Email{stmharry@mit.edu}\\
\Name{Di Jin} \Email{jindi15@mit.edu}\\
\Name{Peter Szolovits} \Email{psz@mit.edu}\\
\addr CSAIL, MIT
}

\editor{}

\begin{document}

\maketitle

\begin{abstract}
Recent developments in Natural Language Processing (NLP) demonstrate that large-scale, self-supervised pre-training can be extremely beneficial for downstream tasks.
These ideas have been adapted to other domains, including the analysis of the amino acid sequences of proteins.
However, to date most attempts on protein sequences rely on direct masked language model style pre-training.
%
In this work, we design a new, adversarial pre-training method for proteins, extending and specializing similar advances in NLP.
We show compelling results in comparison to traditional MLM pre-training, though further development is needed to ensure the gains are worth the significant computational cost.
\end{abstract}

\section{Introduction}

Pre-training, particularly using a self-supervised masked language model (MLM) task over a large corpus, has recently emerged as a powerful tool to improve various prediction and generation tasks, first in natural language processing (NLP) via systems like BERT~\citep{devlin_bert_2019}, and later in other domains, including biomedical domains such as protein sequences\footnote{
Proteins are biological macromolecules responsible for the majority of functions within living cells, represented by, linear chains of ``amino acids,'' over which we perform MLM style pre-training}
~\citep{rao_evaluating_2019,alley_unified_2019,conneau_unsupervised_2019,lu2020self}. However, unlike NLP, where newer methods have brought improved performance, the state of the art for pre-training on protein sequences remains simple MLM style pre-training. For many tasks of interest, this offers only small benefits over prior methods, or even fails to match methods based on human constructed, alignment based features.

In this work, we design a new pre-training method for protein sequences, adapting the traditional MLM task by replacing the random masking scheme with a fully differentiable, adversarial masking model trained to choose which tokens to mask and how in order to make the pre-training model's recovery task \emph{most} difficult subject to a budget constraint. This method is a form of adversarial contrastive estimation~\citep{bose_adversarial_2018}, building on the ideas explored in ELECTRA within NLP~\citep{clark_electra_2019}, but fully differentiable and trained adversarially, a task that is made more feasible given the protein domain's smaller vocabulary. Using an adversarial mask proposal distribution has also been recently explored in connection with reducing gradient variance~\citep{chen2020variance}.

We test our system on the TAPE protein pre-training benchmark system~\citep{rao_evaluating_2019}, achieving modest improvements over comparably trained random pre-training benchmarks, though further development will be needed to ensure this method is worth the increased computational cost. All our code will also be made public after review.

\section{Methods}
\label{sec:model}
Traditionally, to learn a masked language model $\mathcal M_\PT$, we begin with a large, unlabeled corpus of sequences and form training examples by \emph{randomly} choosing a fraction (e.g., 15\% in the case of BERT~\citep{devlin_bert_2019}) of tokens to ``mask.'' The ``masked'' tokens are traditionally noised according to three masking strategies in an 80-10-10 ratio: \texttt{[MASK]}~Masking, in which the masked tokens are replaced with a sentinel, out-of-vocabulary token \texttt{[MASK]}; Keep-original~masking, in which the masked tokens are kept as the original token, but the model is still scored on its ability to form a correct language model prediction for these tokens, and Replace~Masking, in which the masked tokens are replaced with another random valid token from the vocabulary. Given a sentence from the input corpus masked according to this process, the pre-training model $\mathcal M_\PT$ is then tasked to recover the original sentence. 

\begin{figure*}
    \begin{minipage}{0.43\textwidth}
        \centering
        \includegraphics[width=\linewidth]{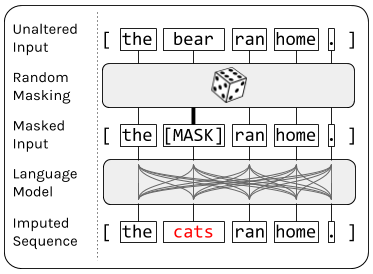}
    \end{minipage}
    \begin{minipage}{0.54\textwidth}
        \centering
        \includegraphics[width=\linewidth]{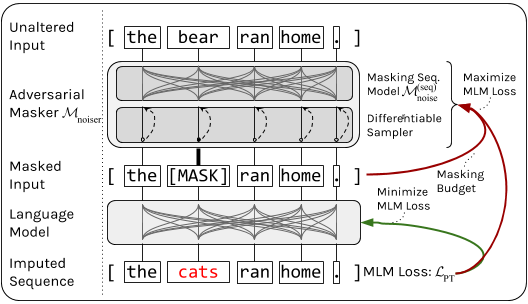}
    \end{minipage}
    \caption{\textit{Left} Traditional random masked language model pre-training. \textit{Right} Our architecture, adversarial contrastive language model pre-training. Note one can use various options for the \emph{Language Model}, \emph{Masking Sequence Model}, or \emph{Differentiable Sampler} components.}
    \label{fig:pre-training_strategies}
\end{figure*}

In this work, we generalize this paradigm by introducing a model $\mathcal M_\NO$ to decide which tokens to mask and how. $\mathcal M_\NO$ is broken down into two parts: a sequence-to-sequence model yielding masking probabilities, and a budgeted differentiable sampling module to actually choose the mask. The whole model is trained end to end via an adversarial approach, such that it learns to mask tokens in a way that makes it most difficult for $\mathcal M_\PT$ to correct. This setup is shown pictorially in Figure~\ref{fig:pre-training_strategies}. Our full adversarial masker $\mathcal M_\NO$ algorithm is given in pseudocode in the Appendix, Algorithms~\ref{alg:noiser},\ref{alg:rss}, Section~\ref{sec:full_methods}, but we will detail several points further here.

\paragraph{Sequence-to-sequence model}
The sequence to sequence model $\mathcal M_\NO^{(\text{seq})}$ (in this work implemented with a gated reucrrent unit (GRU)~\citep{cho_learning_2014} architecture), ingests a sequence of (unmasked) amino acids $\vec x$ and returns a sequence of unnormalized, multidimensional masking scores, $\vec s = \mathcal M_\NO^{(\text{seq})}$. $\vec s$ contains, per-token, a score for (1) whether or not to mask that token at all (we will denote this the \emph{any-mask} score), and (2) a conditional score for each masking option, including \texttt{[MASK]}~masking, keep-original~masking, and replace~masking, with $\vec s$ taking on a score for each possible token that could be used as the replacement mask within the vocabulary (we will refer to these collectively as the \emph{mask-options} scores). These are then passed into our differentiable sampler to obtain a hard, masked sample. 

\paragraph{Differentiable Sampling Options}
We use a slight adaption on the relaxed subset selection algorithm of Sand and Ermon~\citep{xie_reparameterizable_2019} to transform the any-mask scores of $\vec s$ into a vector of normalized probabilities that have two important properties: first, they will average to our masking budget constraint, $\mu$, and, second, they will, in expectation, converge to the their sampled, one-hot approximations (i.e., these probabilities will be a member of a Concrete distribution). This algorithm frames the sampling problem as choosing a fixed-size subset of items from the valid tokens in each sentence, guided by the provided unnormalized scores, through repeated application of the Gumbel-Softmax (GS) trick~\citep{maddison_concrete_2016,jang_categorical_2016}. This process is outlined in pseudo-code in Appendix Algorithm~\ref{alg:rss}.

Next, to decide \emph{how} to mask these tokens, conditioned on the token being masked, we use GS normalization directly on the mask-options scores. Finally, to obtain differentiable one-hot outputs, we use the straight-through estimator~\citep{bengio_estimating_2013}, which simply sets the gradient of the output hard-sample to be equal to the gradient of its source probability, directly, and is especially well suited for probabilities from the Concrete distribution which converge in expectation to their sampled values.

\paragraph{Stabilizing the learning process}
To stabilize the learning process and prevent the masker and MLM model from getting stuck in a local regime of masking, we simply add in additionally a small fraction of random masking, to enable the MLM model to constantly make general progress, which, in turn, forces the masking model to constantly adapt its task to be more difficult than random masking. In order to ensure that some tokens were consistently masked from both varieties, we increased the general masking rate from 15\% total to 20\% total, distributed as 10\% random masking and 10\% adversarial masking in our system.

\paragraph{Overall Training Algorithm}
We then train the system in a traditional adversarial pattern, alternating between several iterations of training the masker to \emph{maximize} the MLM loss, followed by several iterations of training the protein encoder to \emph{minimize} said loss. We use via iterated stochastic gradient descent (using the AdamW optimizer~\citep{loshchilov_decoupled_2018}), with 10 iterations of noiser training followed by 10 iterations of encoder training; these values were determined after a very brief search over possible alternates, using MLM training curve metrics and apparent learning stability to motivate that choice. Additional details about our overall training algorithm are present in supplementary materials Section~\ref{sec:full_methods}.



\paragraph{Data \& Tasks}
We use 4 tasks from the TAPE benchmarking datasets~\citep{rao_evaluating_2019}, profiled in Table~\ref{tab:datasets_summary}. For full details of these datasets and tasks, we encourage readers to refer to~\cite{rao_evaluating_2019}. 

\begin{table}
    \small
    \centering
    \caption{
    A numerical summary of the datasets \& tasks used in this work~\citep{rao_evaluating_2019}.
    }
    \begin{tabular}{llrrrr} \toprule
        Task                & Train & Val. & Test \\\midrule
        Language Modeling   & 32.2M & N/A  & 2.1M   \\
        Secondary Structure & 8678  & 2170 & 513    \\
        Remote Homololgy    & 12312 & 736  & 718    \\
        Fluorescence        & 21446 & 5362 & 27217  \\
        Stability          & 53679 & 2447 & 12839  \\
    \bottomrule \end{tabular}
    \label{tab:datasets_summary}
\end{table}

\paragraph{Experiments}
We compare our adversarial MLM model to a random MLM model (both at 20\% total masking). For our adversarial MLM, $\mathcal M_\NO^{(seq)}$ is a 3-layer, GRU with a 1024 input embedding layer and 512 output layer, and our transformer sequence model is transformer architecture matching the size of that profiled in the TAPE system. Our random MLM system encoder is an identical transformer architecture.

Models were trained on 4 NVIDIA V100 GPUs, ranging in time per model but on the order of roughly 1M encoder iterations at a batch size of 128 for approximately 80\% of training followed by specialization at a larger batch size of 256 using gradient accumulation. Neither memory saving gradients nor mixed-precision training were used. During pre-training, the system was optimized via the AdamW~\citep{loshchilov_decoupled_2018} optimizer with weight decay of of $1e-2$ and learning rate of $1e-4$ following the conventions of TAPE. 

Hyperparameter tuning was performed in a limited, manual manner on train-set MLM results for the pre-training system, and using a grid search on validation set results over batch size, learning rate, weight decays, and early stopping parameters for fine-tuning (using fixed pre-trained models).

\section{Results \& Discussion}
\label{sec:results}
All of our results across all data settings are shown in Table~\ref{tab:final_results}. One can see we obtain improvements over random pre-training on 3 of 4 tasks, though in all cases changes are mild. Given the increased computational cost of this style of training (approximately 2x due to masker and encoder training), these gains are likely not currently worth these minor improvements. However, they do establish that this direction of research may be a viable vehicle to improve protein pre-training, with further improvements. Several directions stand out to offer such improvements. Firstly, we believe it is likely that higher-capacity noising models would work better, though early attempts with this architecture proved too unstable to train effectively. Second, we feel we could more effectively train this system with larger batch sizes (which has been shown to offer improvements in other pre-training contexts) through the use of mixed-precision training and gradient checkpointing. Finally, the incorporation of an importance sampling re-weighting penalty on the learning objective, rather than use of partial random masking, as in the style of \citep{chen2020variance} may offer further improvements here.

\begin{table}
    \centering
    \small
    \caption{Final results for a random MLM and our adversarial MLM, reported in accuracy / amino acid for secondary structure (SS), accuracy / sequence for remote homology (RH), and Spearman correlation coefficient for fluorescence and stability. Variance is sourced over repeated FT runs only, not PT runs, as the latter would be computationally intractable.}
    \begin{tabular}{lrr}
    \toprule
        Task         & Random                 & Adv. (Ours)            \\ \midrule
        S. S.        & $72.1 \pm 0.1\%$       & $\rr{72.7 \pm 0.2\%}$  \\
        R.H.         & $19.2 \pm 0.8\%$       & $\rr{20.0 \pm 2.0\%}$  \\
        Fluorescence & $\rr{0.678 \pm 0.002}$ & $0.677 \pm 0.002$      \\
        Stability    & $0.626 \pm 0.044$      & $\rr{0.630 \pm 0.009}$ \\
    \bottomrule
    \end{tabular}
    \label{tab:final_results}
\end{table}

\section{Conclusion}
In this work, we design a novel adversarial, contrastive contextual embedding system for protein sequences, attaining improvements over comparable random pre-training runs on three of four tasks in the TAPE benchmark, though these improvements are minor and further work will be needed to ensure realized gains are worth the increased computational cost.


\bibliography{references}

\clearpage
\appendix
\onecolumn

\section{Full Methods}
\label{sec:full_methods}
Pseudocode algorithms for the full masker, the straight through sampler, and the full adversarial system, including masker and protein encoder, are shown in the below Algorithms:

\begin{algorithm}
    \SetAlgoLined
    \DontPrintSemicolon
    \SetKwFunction{rssSampler}{rss\_sampler}
    \SetKwProg{proc}{Procedure}{}{}
    \SetKwFunction{softmax}{softmax}
    \SetKwFunction{round}{round}
    \SetKwFunction{uniform}{Uniform}
    \proc{\rssSampler{$\vec s$, \texttt{valid\_tokens\_mask}, $\rho, t$}}{
        \KwResult{$\vec p$}
        \KwIn{$\vec s$: Per-element selection scores, \texttt{valid\_tokens}: Binary mask highlighting valid inputs within the batch, $\rho$: Desired masking fraction, $t > 0$: sampling temperature value.}
        \; \\
        Initialize $\varepsilon \gets 10^{-18}$, $\vec y_\text{soft} \gets \vec 0$, $\vec g \gets \vec s - \log(-\log(\uniform{0, 1}))$ \; \\
        \texttt{seq\_lens} $\gets$ \texttt{valid\_tokens\_mask.sum(axis=1)}\; \\
        \texttt{subset\_sizes} $\gets$ \round{\texttt{seq\_lens * }$\rho$} \; \\
        \texttt{subset\_sizes\_left} $\gets$ \texttt{subset\_sizes.expand\_as(valid\_tokens\_mask)} \; \\
        \; \\
        \While{\texttt{subset\_sizes\_left} $> 0$}{
            \texttt{khot\_mask} $\gets \max((1-\vec y_\text{soft})\texttt{ * valid\_tokens\_mask}, \varepsilon)$\; \\
            $\vec g \texttt{ += } \log(\texttt{khot\_mask}))$ \; \\
            $\vec y_\text{soft}$\texttt{ += }\softmax{$\vec g / t,\, \texttt{dim}=-1$}\; \\
            \; \\
            \texttt{subset\_sizes\_left -= 1}\; \\
            \texttt{valid\_tokens\_mask *= (subset\_sizes\_left > 0).float()}\; \\
        }
    
        \KwRet $\vec y_\text{soft}$ \;
    }
    \caption{Our variable masking rate per-batch variant on the relaxed subset selection algorithm of~\citep{xie_reparameterizable_2019}}
    \label{alg:rss}
\end{algorithm}

\begin{algorithm}
    \SetAlgoLined
    \DontPrintSemicolon
    \SetKwFunction{rssSampler}{rss\_sampler}
    \SetKwFunction{runNoiser}{run\_noiser}
    \SetKwFunction{runEncoder}{run\_encoder}
    \SetKwFunction{updateNoiser}{update\_noiser}
    \SetKwFunction{updateEncoder}{update\_encoder}
    \SetKwFunction{softmax}{softmax}
    \SetKwFunction{uniform}{Uniform}
    \SetKwFunction{straightThrough}{straight\_through}

    \SetKwProg{proc}{Procedure}{}{}
    \KwIn{$\rho$: masking rate; $t$: mask option temperature; \texttt{mask\_id}: \texttt{[MASK]} token ID in vocabulary
    }
    
    \proc{\runNoiser{$\vec \theta, \vec x, \texttt{valid\_tokens\_mask}$}}{
        $\vec s \gets \mathcal M_\NO^{(\text{seq})}(\vec x, \texttt{valid\_tokens\_mask}; \vec \theta)$ \; \\
        $\vec p_{\text{mask overall}} \gets$ \rssSampler{$\vec s\texttt{[:,:,0]}, \texttt{valid\_tokens\_mask}, \rho, t$}\; \\
        $\vec g_{\text{mask type}} \gets \vec s\texttt{[:,:,1:]} - \log(-\log(\uniform{0, 1}))$ \; \\
        $\vec p_{\text{mask type}} \gets $\softmax{$\vec g_{\text{mask type}}/t$} \; \\
        \; \\
        $\tilde{\vec x} \gets$ \straightThrough{$\vec x, \vec p_{\text{mask overall}}, \vec p_{\text{mask type}}$}\; \\
        \KwRet $\tilde{\vec x}$ \;
    }
    \caption{The full noising process, taking in the raw, un-masked input vector $\vec x$ and producing a noised version of the input suitable for pre-training an MLM. \texttt{rss\_sampler} is shown in Algorithm~\ref{alg:rss} and \texttt{straight\_through} is shown in the supplementary materials, Algorithm~\ref{alg:straightThrough}}
    \label{alg:noiser}
\end{algorithm}

\begin{algorithm}
    \SetAlgoLined
    \DontPrintSemicolon
    \SetKwFunction{straightThrough}{straight\_through}
    \SetKwProg{proc}{Procedure}{}{}
    \SetKwFunction{argmaxMask}{argmax\_mask}
    \SetKwFunction{oneHot}{one\_hot}
    \SetKwFunction{where}{where}
    \SetKwFunction{detach}{detach}
    \KwResult{$\tilde{\vec x}$}
    \KwIn{$\vec x$: one-hot encoding of the input, which can be multiplied by an embedding layer to produce distributed embeddings of each token. $\vec p_{\text{mask overall}}$: probabilities of masking each token in any form; $\vec p_{\text{mask type}}$: probabilities of each kind of masking of each token, or \texttt{None} for simple masking; \texttt{mask\_id}: \texttt{[MASK]} token ID in vocabulary; $v_\text{idx}$: the start index of the vocabulary, after skipping past all control tokens (e.g., \texttt{[MASK]}).}
    \proc{\straightThrough{$\vec x$, $\vec p_{\text{mask overall}}$, $\vec p_{\text{mask type}}$}}{
        
        \texttt{mask\_ANY} $\gets$ \where{$\vec p_{\text{mask overall}} > 0.5$, $\vec 1$, $\vec 0$}\; \\
        $\vec x_{\text{masked}} \gets \vec 0$\; \\
        \eIf{$\vec p_{\text{mask type}} =$ \texttt{None}}{
            $\vec x_{\text{masked}}$\texttt{ += }\oneHot{\texttt{mask\_id}} $+ p_{\text{mask overall}} - $\detach{$p_{\text{mask overall}}$} \; \\
        }{
            $\vec M \gets$ \argmaxMask{$\vec p_{\text{mask type}}$} $+ p_{\text{mask type}} -$ \detach{$p_{\text{mask type}}$}\; \\
            \; \\
            $\vec m_{\text{\texttt{[MASK]}}} \gets \vec M\texttt{[:,:,0]}$ \; \\
            $\vec m_{\text{keep}} \gets \vec M\texttt{[:,:,1]}$ \; \\
            $\vec m_{\text{replace}} \gets \vec M\texttt{[:,:,2:]}$ \; \\
            \; \\
            $\vec x_{\text{masked}}$\texttt{ += }\oneHot{\texttt{mask\_id}}\texttt{ * }$\vec m_{\texttt{[MASK]}}$ \; \\
            $\vec x_{\text{masked}}$\texttt{ += }$\vec x\texttt{ * }\vec m_{\text{keep}}$ \; \\
            $\vec x_{\text{masked}}\texttt{[:,:,}v_\text{idx}\texttt{:]}$\texttt{ += }$\vec m_{\text{replace}}$ \; \\
        }
        
        $\tilde{\vec x} \gets (1 - \texttt{mask\_ANY})\vec x + (\texttt{mask\_ANY})\vec x_{\text{masked}}$\;

        \KwRet $\tilde{\vec x}$ \;
    }
    \caption{Straight through sampler, accounting for all three modes of masking.}
    \label{alg:straightThrough}
\end{algorithm}

\begin{algorithm}
    \SetAlgoLined
    \DontPrintSemicolon
    \SetKwFunction{rssSampler}{rss\_sampler}
    \SetKwFunction{runNoiser}{run\_noiser}
    \SetKwFunction{runEncoder}{run\_encoder}
    \SetKwFunction{updateNoiser}{update\_noiser}
    \SetKwFunction{updateEncoder}{update\_encoder}
    \SetKwFunction{softmax}{softmax}
    \SetKwFunction{uniform}{Uniform}
    \SetKwFunction{sgd}{SGD}
    \SetKwProg{proc}{Procedure}{}{}

    \proc{\runEncoder{$\vec \theta, \vec \phi, \vec x, \texttt{valid\_tokens\_mask}$}}{
        $\tilde{\vec x} \gets$ \runNoiser{$\vec \theta, \vec x$, \texttt{valid\_tokens\_mask}}\; \\
        $\vec p_\text{reconst.} \gets \mathcal M_\PT(\tilde{\vec x}, \texttt{valid\_tokens\_mask}; \vec \phi_\PT)$\; \\
        \KwRet $\mathcal L(\vec p_\text{reconst.}, \vec x, \texttt{valid\_tokens\_mask})$\; \\
    }
    
    \proc{\updateNoiser{$\vec \theta, \vec \phi, \vec x, \texttt{valid\_tokens\_mask}$}}{
        \KwRet\sgd{-\runEncoder{$\vec \theta, \vec \phi, \vec x, \texttt{valid\_tokens\_mask}$}}
    }
    \proc{\updateEncoder{$\vec \theta, \vec \phi, \vec x, \texttt{valid\_tokens\_mask}$}}{
        \KwRet \sgd{\runEncoder{$\vec \theta, \vec \phi, \vec x, \texttt{valid\_tokens\_mask}$}}
    }
    \caption{Noiser \& Encoder Update Steps. In practice we use more the more advanced AdamW SGD variant.}
    \label{alg:updates}
\end{algorithm}

\begin{algorithm}
    \SetAlgoLined
    \DontPrintSemicolon
    \SetKwFunction{updateNoiser}{update\_noiser}
    \SetKwFunction{updateEncoder}{update\_encoder}
    \SetKwFunction{getBatch}{get\_batch}
    \KwResult{$\vec \theta^*$, $\vec \phi_\PT^*$}
    \KwIn{$n_\text{noiser}$: \# of noiser iterations per cycle}
    \KwIn{$n_\text{encoder}$: \# of encoder iterations per cycle}
    Initialize $\vec \theta^{(0)}, \vec \phi_\PT^{(0)}$ randomly\;
    Initialize $i \gets 1$, \texttt{mode} $\gets$ \texttt{PRE\_TRAINING\_NOISING}\; \\
    \While{Not Converged}{
        $\vec x, \texttt{valid\_tokens\_mask} \gets $\getBatch{$i$} \; \\
        \eIf{\texttt{mode} $=$ \texttt{PRE\_TRAINING\_NOISING}}{
            $\vec \theta^{(i+1)} \gets $ \updateNoiser{$\vec \theta^{(i)}, \vec \phi_\PT^{(i)}, \vec x, \texttt{valid\_tokens\_mask}$} \; \\
            \If{$i \bmod n_\text{noiser} = 0$}{
                \texttt{mode} $\gets$ \texttt{PRE\_TRAINING\_ENCODING}\;
            }
        }{
            $\vec \phi_\PT^{(i+1)} \gets $ \updateEncoder{$\vec \theta^{(i+1)}, \vec \phi_\PT^{(i)}, \vec x, \texttt{valid\_tokens\_mask}$} \; \\
            \If{$i \bmod n_\text{encoder} = 0$}{
                \texttt{mode} $\gets$ \texttt{PRE\_TRAINING\_NOISING}\;
            }
        }
        $i \gets i+1$ \; \\
    }
    \KwRet $\vec \theta^{(i)}, \vec \phi_\PT^{(i)}$ \;
    \caption{Our adversarial contrastive training algorithm}
    \label{alg:adv_training}
\end{algorithm}
\end{document}